\pdfoutput=1

\newcommand{\squishlist}{
\begin{list}{$\bullet$}
{   \setlength{\itemsep}{0pt}
   \setlength{\parsep}{3pt}
   \setlength{\topsep}{3pt}
   \setlength{\partopsep}{0pt}
   \setlength{\leftmargin}{1.5em}
   \setlength{\labelwidth}{1em}
   \setlength{\labelsep}{0.5em} } }
\newcounter{Lcount}
\newcommand{\squishlisttwo}{
\begin{list}{\arabic{Lcount}. }
  { \usecounter{Lcount}
 \setlength{\itemsep}{0pt}
 \setlength{\parsep}{0pt}
 \setlength{\topsep}{0pt}
 \setlength{\partopsep}{0pt}
 \setlength{\leftmargin}{2em}
 \setlength{\labelwidth}{1.5em}
 \setlength{\labelsep}{0.5em} } }
\newcommand{\squishend}{\end{list} }

\documentclass[11pt]{article}
\usepackage{graphicx}
\usepackage{amsmath}
\usepackage{listings}
\usepackage{xcolor}

\usepackage[preprint]{acl}

\usepackage{times}
\usepackage{latexsym}

\usepackage[T1]{fontenc}

\usepackage[utf8]{inputenc}

\usepackage{microtype}

\usepackage{inconsolata}
\usepackage{float}
\usepackage{graphicx}

\title{Cluster Purge Loss: Structuring Transformer Embeddings for Equivalent Mutants Detection}

\author{
    Adelaide Danilov \and Aria Nourbakhsh \and Christoph Schommer \\
    University of Luxembourg \\ Esch-sur-Alzette, Luxembourg}

\begin{document}
\maketitle
\begin{abstract}
Recent pre-trained transformer models achieve superior performance in various code processing objectives. However, although effective at optimizing decision boundaries, common approaches for fine-tuning them for downstream classification tasks — distance-based methods or training an additional classification head — often fail to thoroughly structure the embedding space to reflect nuanced intra-class semantic relationships. Equivalent code mutant detection is one of these tasks, where the quality of the embedding space is crucial to the performance of the models. We introduce a novel framework that integrates cross-entropy loss with a deep metric learning objective, termed Cluster Purge Loss. This objective, unlike conventional approaches, concentrates on adjusting fine-grained differences within each class, encouraging the separation of instances based on semantical equivalency to the class center using dynamically adjusted borders. Employing UniXCoder as the base model, our approach demonstrates state-of-the-art performance in the domain of equivalent mutant detection and produces a more interpretable embedding space.
\end{abstract}

\section{Introduction}
Since the introduction of the transformer architecture \citep{vaswani_attention_2017}, large language models showed radical improvements on a great range of NLP \citep{raiaan_review_2024} and code-related \citep{zheng_survey_2023} tasks. Moreover, LLMs can be further fine-tuned for downstream tasks using a domain-specific dataset\citep{parthasarathy_ultimate_2024}. For bi-directional encoder transformers — ones excelling in analyzing existing code \citep{nijkamp2023codegen2lessonstrainingllms} — the standard approach in fine-tuning is to use a task-specific head to train with the rest of the transformer. However, for some such tasks requiring deep semantic understanding, the structure of the resulting embedding space is extremely important \citep{li_diffusion-lm_2024}, and the method above may struggle to provide it adequately. One such task is equivalent mutant detection (EMD).

\textit{Mutation testing} \citep{jia_analysis_2011} is a software testing approach. The principle of this approach is to generate programs based on an initial program under test by applying mutation operators. Such generated programs called \textit{mutants}, are supposed to exhibit altered behavior so they can be used to examine the adequacy of test suites for that program. A mutant passing some test cases in a suite signifies the inability to catch a potential bug. \autoref{sec:mutesting_appendix} presents a graphic explanation and \autoref{sec:examples_appendix} shows examples of generated mutants. Although mutation testing is widely known and has applications in other fields of testing (e.g., test case prioritization \citep{7381798}, bug detection \citep{10.1145/3276517}, localization of faults \citep{10.1002/stvr.1509}), one of the main reasons hindering its adoption is the existence of \textit{equivalent mutants}. Such programs are semantically equal to their origin, thus producing the same output. Equivalent mutants have posed a persistent challenge, as their presence distorts test outcomes and the \textit{mutation score}, which makes their detection necessary. Hereinafter, we will understand \textit{equivalent mutants} as mutants semantically equivalent to their origin program and \textit{non-equivalent mutants} as those semantically different to it.

The history of EMD includes a considerable number of different approaches such as constraint-based testing \citep{baer_mutantdistiller:_2020}, compiler optimizations \citep{7194639, kintis_detecting_2018} and machine learning, i.e. SVM \citep{https://doi.org/10.1002/smr.2238} and RNN based approaches \citep{peacock_automatic_2021}. A recent study by \citet{tian_large_2024} showed that LLMs significantly outperform previous techniques' Precision, Recall, and F1-score, demonstrating an average 35.69\% gain in the latter. Their approach achieved the highest values with BERT \citep{devlin_bert:_2019}-based uniXCoder \citep{guo_unixcoder:_2022}, utilizing graph-guided masked attention(GGMA) based on the representation of dependency relations between variables in the source code - Data Flow\citep{guo_graphcodebert:_2020}.

We hypothesize that even though mutants descended from the same original program - \textit{mutant class} - are clustered and separated in the embedding space from other classes, the subtle intra-class differences between \textit{equivalent} and \textit{non-equivalent} mutants are not adequately formed and captured by the fine-tuned LLMs and the classifier alone. A further hypothesis was put forward that such properties can be obtained by utilizing deep metric learning (DML) \citep{MOHAN202359} and, in turn, improve the classification of mutants. However, most DML approaches such as contrastive loss \citep{contrastive}, triplet loss \citep{DBLP:conf/cvpr/SchroffKP15}, proxyNCA++ \citep{teh_proxynca++:_2020} concentrate at the inter-class level, without explicitly structuring instances inside formed class clusters. 

In our work, we confirm the hypothesis about the embedding space and propose an approach of carefully combining Cross-Entropy Loss from the classification head with a new loss function named \textit{Cluster Purge Loss}. The idea of this function is that for each class, we update the Exponential Moving Average of all distances between equivalent mutants and their origin, do the same for non-equivalent mutants, and then try to push or pull mutants beyond the resulting average radius of their counterparts just enough to aid fine-tuning with distinguishing between them.

By conducting an ablation study, using the same LLM (uniXCoder), classifier head (RoBERTa classifier), Java mutant dataset, number of epochs, batch size and optimizer hyperparameters, we show that our method increases \textbf{precision(5.12 pp), recall(0.57 pp) and F1-score(2.24 pp)} compared to the highest results obtained by \citet{tian_large_2024}.
Also, we made a comparison with the DML baseline - the setup where Cross-Entropy Loss was combined with the contrastive loss instead of the proposed CPL, finding that CPL outperforms it with \textbf{1.28 pp} F1-score increase. Lastly, we repeated the comparisons above on a dataset of mutants written in C, showing that CPL is generalizable and outperforms others in this setting as well.

Thus, the contributions of this paper can be summarized as follows:
\squishlist
\item Introduced a new Deep Metric Learning loss function, which aims not to organize classes of instances but to adjust semantic relationships inside each already formed cluster according to the given binary distinction.
\item Showed that applying DML approach can be beneficial during fine-tuning a large language model for specific downstream tasks.
\item Obtained results superior to SOTA in EMD while isolating the performance gains attributable to the proposed approach.
\squishend
\begin{figure}[h]
    \centering
    \includegraphics[width=1\linewidth]{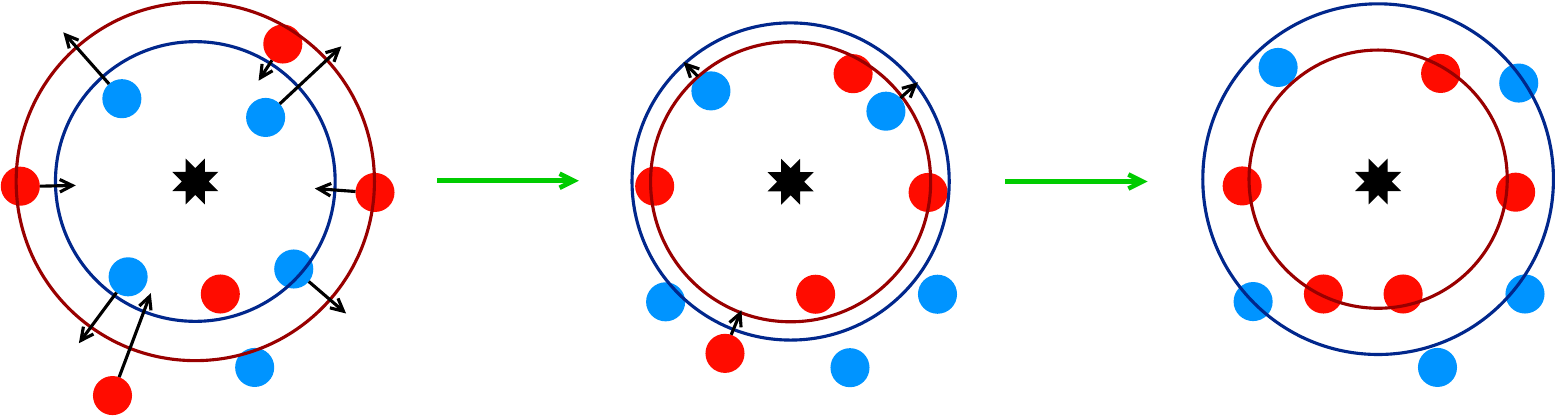}
    \caption{CPL conditions mutants to cross the Exponential Moving Average boundary of their counterparts, depicted as a circle of the opposite color, which is adjusted throughout training.}
    \vspace{-5mm}
    \label{fig:enter-label}
\end{figure}
\section{Proposed approach}

\autoref{fig:baseline_embeddings} illustrates a 2D t-SNE visualization of baseline model mutant embeddings of two classes generated by samples 1408 and 2001 from the Java mutant dataset. In both cases, the distributions of equivalent and non-equivalent mutants within each class overlap significantly, and clustering fails to reflect semantic relationships. We hypothesize that introducing a secondary loss function explicitly designed to differentiate mutants in the embedding space based on their semantic equivalence to the ancestor program can facilitate the emergence of a more organized embedding space during fine-tuning. This improved organization, in turn, may enhance the performance of the classifier head by making distinctions between instances more straightforward.

The joint loss function is formulated as follows:
\begin{equation}
    L = L_{\text{CPL}} \cdot \lambda + L_{\text{CE}}
\label{eq:1}
\end{equation}
Where $ L_{\text{CE}}$ is Cross-Entropy Loss obtained from the classifier head and $L_{\text{CPL}}$ is the proposed Cluster Purge Loss. Combining loss functions may lead to a situation where their goals may be inconsistent \citep{9025455}. To combat this and balance the weight of each loss, the hyperparameter $\lambda$ is used.
\subsection{Cluster Purge Loss}\label{sec:2.1}
We formalize the problem. Assume the minibatch size of m where each sample is:
 \begin{equation}
    (k_{i}, o_{k_i}, s_{i}, l_{i})\text{ where }l_i \in \{0,1\}, i \in \{0, \dots, m\}
\label{eq:2}
\end{equation}
$k_i$ is the unique identifier of a class, $o_{k_i}$ is the embedding of the original program associated with $k_i$, $s_{i}$ is the embedding of the another mutant belonging to $k_i$, and $l_{i}$ represents its equivalence to the origin.
Select all unique classes in the minibatch:
\begin{equation}
    K = \{ k_j \mid j \in \{0, \dots, m\}\}
\end{equation}
For each unique class, find distances between its origin and equivalent mutants in the minibatch, where $d_c^+$ is a tuple of such distances for class $c$:
\begin{equation}
    \begin{split}
    \forall c \in  K,d_c^+ =\ &(\mathrm{dist}(o_{k_i},s_{i}) \mid k_i = c \land l_i = 1, \\ &i \in \{0, \dots, m\})  
    \end{split}
\end{equation}
The equation for Exponential Moving Average, where $\gamma$ is a \textit{smoothing factor}: 
\begin{equation}
    \text{EMA}_{n+1} = \text{EMA}_{n} \cdot (1 - s) + x \cdot s,\ s = \frac{2}{\gamma +1}
\end{equation}
Derive closed form for several $x_1,...,x_h$:
\begin{equation}
    \text{EMA}_{n+h} = \text{EMA}_{n} \cdot (1 - s)^h + s\cdot\sum_{j=1}^{h}x_j \cdot (1- s)^{h-j}
\end{equation}
Using eq.6 we can update EMA of distances from the origin to equivalent mutants for each class encountered in the minibatch. The resulting average for the class $c$ we will call a \textit{positive verge} $v_c^+$. If $v_c^+$ is updated for the first time, then it is pre-initialized with the ${d_c^+}_0$. Formulated as the following:
\begin{equation}
    \forall c \in  K,v_c^+ = 0 \implies v_c^+ = {d_c^+}_0
\end{equation}
\begin{equation}
    \begin{split}
    \forall c \in  K,v_c^+ &\gets \  v_c^+ \cdot (1 - s)^{\lvert d_c^+ \rvert} + \\ & s\cdot\sum_{j=1}^{{\lvert d_c^+ \rvert}}{d_c^+}_j \cdot (1- s)^{{\lvert d_c^+ \rvert}-j},\ s=\frac{2}{\gamma +1}
    \end{split}
\end{equation}
Next, we carry out the same calculations for non-equivalent mutants to find a tuple of distances $d_c^-$(for samples with $l_i = 0$) and a \textit{negative verge} $v_c^-$:

Finally, we can compute the loss function based on the current minibatch:
\begin{equation}
    \begin{aligned}
    L_{\text{CPL}} =\ &\frac{1}{m}\sum_{i=1}^{m}(\left[\mathrm{dist}(o_{k_i},s_{i}) - v_{k_i}^- + \zeta\right]_+^\alpha \cdot l_i + \\ &\left[v_{k_i}^+ -\mathrm{dist}(o_{k_i},s_{i}) +  \zeta\right]_+^\beta \cdot (1 - l_i))
    \end{aligned}
\end{equation}
If $l_i = 1$, then $s_i$ is equivalent and the calculation is as follows: distance from $s_i$ to the origin $o_{k_i}$ of its class $k_i$ minus the negative verge for $k_i$ and plus the margin $\zeta$; then ReLU is applied and the resulting expression is raised to the power of $\alpha$. Such formulation encourages keeping the distance of equivalent mutants to the origin less than the boundaries of non-equivalent mutants by $\zeta$. The same principle applies if $s_i$ is non-equivalent, but in the opposite direction and with the positive verge.

Hyperparameters $\alpha$ and $\beta$ are introduced to control growth of the loss function for both cases separately when asymmetric structuring is beneficial.
\section{Experimentation}
To assess our approach we conducted an ablation study. UniXCoder(110M) fine-tuned on Java mutant dataset with GGMA and cross-entropy loss, which was found by \citet{tian_large_2024} to perform the best in terms of F1-score among LLMs and other approaches, was taken as the baseline of the study. It was compared with UniXCoder fine-tuned with the same setup modified to use a sum of cross-entropy and CPL. The altered setup inherited the values of all shared hyperparams from the baseline.

In addition, we made a comparison against a combination of cross-entropy with adapted contrastive loss, which is described in \autoref{sec:experiments_appendix}.

Lastly, to investigate generalization of CPL, experiments listed in the main text and \autoref{sec:experiments_appendix}, which were carried out on the primary Java dataset, were repeated for the mutant dataset in C language and reported separately in \autoref{sec:generalization_appendix}.

\subsection{Dataset preparation}\label{sec:3.1}
The dataset utilized in the baseline study was derived from MutantBench \citep{noauthor_mutantbench:_2021} aggregating many previously published datasets. \citet{tian_large_2024} preprocessed it and obtained 3302 pairs of Java mutants with the same origin method, accompanied by the equivalency label. ${train}_{base}$ of size 1652 was constructed by sampling 50/50 split of equivalent and non-equivalent mutant pairs, while ${test}_{base}$ was created with the remaining 1650 mutants.

During preprocessing, we determined the origins of all mutants in the datasets and, based on them, introduced 52 mutant classes, each assigned a sequential id. To construct ${train}_{cpl}$, we augmented each pair of mutants from ${train}_{base}$ with the class id, resulting in 1590 pairs. The same operation was performed to create ${test}_{cpl}$ with 1580 pairs. The number of pairs in the obtained Java datasets is slightly lower due to duplicates being removed.

\subsection{Implementation} \label{sec:3.2}
${model}_{base}$ is the pre-trained UniXCoder paired with the RoBERTa classification head. During fine-tuning the input sequence is constructed from the source code of each mutant and Data Flow Graph. The input sequence is converted into input vectors, following \citet{guo_graphcodebert:_2020}, and fed to the forward pass method. Input token embeddings and GGMA matrix are calculated and passed to the UniXCoder in the encoder mode.The embedding representing a mutant is acquired by taking a normalized CLS token out of the last layer output. Subsequently, pairs of embeddings are passed to the classifier head and softmax. Finally, the cross-entropy loss is computed using equivalence probability labels.

For ${model}_{cpl}$, we set up the similar pipeline following Guo et al. and Tian et al. Features are extracted taking into account the addition of class id in ${train}_{cpl}$, and ${model}_{cpl}$ is implemented to include the calculation of CPL. The Model class stores verges in the buffer during fine-tuning, preserving them between epochs, and computes CPL and the final loss as described in \autoref{sec:2.1}.

\subsection{Evaluation} \label{sec:3.3}
\subsubsection{Experiment results}
We conducted 56 experiments by fine-tuning ${model}_{cpl}$ on ${train}_{cpl}$ and evaluating on ${test}_{cpl}$ with $dist$ being normalized cosine distance, $\gamma = 12$, $\alpha = 2$, $\beta = 1/2$, $\zeta \in [-0.06, 0.01]$ with a step 0.01 and $\lambda \in [1.00, 1.30]$ with a step 0.05. The rationale for choosing hyperparameters is present in \autoref{sec:hyperparam_appendix}. The obtained metrics were compared against ${model}_{base}$ fine-tuned with the same number of epochs = 30, batches = 4 and other shared hyperparameters. The results for all combinations of $\lambda$ and $\zeta$ are presented in \autoref{fig:matrix} where acquired precision(P), recall(R) and F1-score(F1) are stated. The best result in terms of F1-score is (P: 95.31\%, R: 85.41\%, F1: 89.46\%) at $\lambda$ = 1.15 and $\zeta$ = -0.05. Given that the metrics of ${model}_{base}$ is (P: 90.19\%, R: 84.84\%, F1: 87.22\%), the absolute gain is (P: \textbf{5.12 pp}, R: \textbf{0.57 pp}, F1: \textbf{2.24 pp}). In \autoref{tab:comparison} we also include results obtained for ${model}_{base}$ by Tian et al. using epochs = 10 and ${model_{contrast}}$ described in \autoref{sec:experiments_appendix}. It is clear that our approach shows the best results for all metrics.

\begin{table}[t]
    \centering
    \resizebox{0.9\columnwidth}{!}{
      \begin{tabular}{|lccc|}
         \hline
         \textbf{Technique} & \textbf{Precision} & \textbf{Recall} & \textbf{F1-score} \\
         \hline\hline
         ${model}_{CPL}$ & \textbf{95.31\%} & \textbf{85.41\%} & \textbf{89.46\%} \\
         ${model}_{contrast}$ & 92.75\%, & 84.84\% & 88.18\% \\
         ${model}_{base}$ & 90.19\% & 84.84\% & 87.22\%  \\
         ${model}_{base}$, Tian & 94.33\% & 81.81\% & 86.58\%  \\
         \hline
      \end{tabular}
    }
    \caption{Comparison with the baselines}
    \label{tab:comparison}
    \vspace{-5mm}
\end{table}

\subsubsection{Impact on embeddings distribution}
To prove the hypothesis about Cluster Purge Loss promoting more organized embedding space, which is beneficial for EMD, the embeddings of mutants with origin 1408 and 2001 were extracted from the best performing ${model}_{cpl}$ and plotted after applying T-SNE(\autoref{fig:embeddings_cpl}). Non-equivalent mutants can be observed to be distributed significantly further away from the origin, while the distance to equivalent mutants varies. For the origin 1408, 2 clusters of equivalent mutants were formed, the first one is close to the origin, while the second is distanced from it. The latter can be explained by the negative $\zeta$ as discussed in \autoref{sec:hyperparam_appendix}.

However, T-SNE doesn't always preserve global structure well. To investigate observations, the mean distance of embeddings of all non-equivalent mutants to their origin was computed: $0.105 \pm 0.133$ for ${model}_{base}$ and $0.398 \pm 0.303$ for ${model}_{cpl}$ with p < 0.0001. For all equivalent mutants: $0.111 \pm 0.215$ for ${model}_{base}$ and $0.189 \pm 0.284$ for ${model}_{cpl}$ with p = 0.83. That means that the ratio between the mean distance of non-equivalents to the origin and the mean distance of equivalents to the origin increased from 0.95 to 2.11 and is attributed to the statistically significant change in the distribution of the non-equivalent mutants.

Thus, we can conclude that our hypothesis holds and the introduction of CPL improved the performance on the equivalent mutant detection task by promoting the semantic meaning on distances between embeddings in the intra-class context.

\section{Conclusion}
In this study we introduced new Deep Metric Learning loss function named Cluster Purge Loss which organizes instances in already formed class clusters based on the semantical similarity to the class center. By the ablation study, we showed that using CPL in the joint loss formulation with the cross-entropy loss shows state-of-the-art performance in equivalent mutant detection and found out that it is attributed to CPL impact on the embedding space.

\section{Limitations}
The main limitation of our work is that we ran only one trial for each of the 196 hyperparameter experiments due to limited computational resources. Conducting multiple runs for each experiment would help reduce variance caused by randomness and produce more robust conclusions.
\bibliography{custom}

\appendix

\section{Hyperparameters selection}
\label{sec:hyperparam_appendix}
To evaluate our approach, we conducted a series of experiments aiming to explore the hyperparameter space of Cluster Purge Loss. As the distance function normalized cosine distance was chosen:
\begin{equation}
    \begin{split}
    \text{dist}(a,b) = 1 - \frac{\text{cossim}(a,b) + 1}{2} 
    \end{split}
\end{equation}
Inverse formulation means that the codomain is [0,1] where 0 indicates collinearity of vectors. The smoothing factor $\gamma$ was chosen as 12 based on the preliminary experiments. The value of the exponent of a loss term for equivalent mutants $\alpha$ is 2 and the exponent of a loss term for non-equivalent mutants $\beta$ is 1/2.
Such initial values of $\alpha$ and $\beta$ are based on the assumption that equivalent mutants are already located close enough to their origin, and to give semantic similarity properties to the embedding space, emphasis must be placed on changing the distribution of non-equivalent mutants. Since the square function shows sublinear growth on values close to 0 included in the codomain [0,1] of the distance function, and the root function, on the contrary, grows superlinearly, then the loss value for non-equivalent mutants will grow faster with the distance from the verge than for equivalent ones thereby achieving the desired goal. 

The margin $\zeta$ between mutants and the corresponding verges and the coefficient $\lambda$ at $L_{CPL}$ are considered the most influential and explored in ranges: $\zeta \in [-0.06, 0.01]$ with a step 0.01 and $\lambda \in [1.00, 1.30]$ with the step 0.05. Such intervals are chosen based on the preliminary findings showing that smaller values are more favorable. For $\lambda$ we explain it by the assumption that Cluster Purge Loss is more beneficial in the setup as the lesser term in the equation shifts the negative gradient towards a more optimal solution by imposing the semantic meaning on the distance. It follows that $\lambda$ tends to be around 1 since initially, the value of ${L}_{CPL}$ is several times less than that of ${L}_{CE}$. For $\zeta$, we assume that since the model’s ability to capture semantic differences between mutants is imperfect, a negative boundary can create an “error zone” for those mutants that cannot be correctly ordered without worsening the arrangement of the rest.
\section {Experiments on integrating other DML approaches}
\label{sec:experiments_appendix}
In addition to the comparison with the baseline derived from \citet{tian_large_2024}, we also considered triplet loss and contrastive loss as standard DML baselines. However, preliminary experiments with triplet loss showed that generating all possible triplets per each mutant class and constructing the dataset with up to Java $10^4$ triplets via stratified sampling demonstrates F1-score no higher than 
only \textbf{64.37\%}. Since further gains in performance would require either increasing the size of the dataset, which would be prohibitive in terms of the computational cost, or exploring a triplet mining strategy, it was decided to favor contrastive loss as the DML baseline.  

We conducted 42 experiments on the primary Java dataset where ${L}_{contrast}$ was used instead of ${L}_{CPL}$ in Equation
\eqref{eq:1}. The purpose of these experiments is to investigate how significant is the impact of contrastive loss on the properties of the embedding space in terms of equivalent mutant detection and to compare Cluster Purge Loss with it. The train setup inherited all shared hyperparameters from the baseline in the same manner as the CPL one did.
\subsection{Contrastive loss}
The principle of the contrastive loss is a direct application of the deep metric learning approach - to learn such a representation of the embedding space that similar samples lie close to each other while dissimilar is pushed apart.
Specifically, it minimizes the distance between pairs of similar instances and maximizes it between pairs of the dissimilar ones.
In the scope of the Equivalent Mutant Detection problem, adjusting distances between different classes of mutants is unnecessary, as the nature of the problem lies in distinguishing equivalent and non-equivalent mutants inside each class separately. Further, within each class, we need to consider only pairs consisting of a mutant and its origin since we are interested in regulating the distance of the two types of mutants to the said origin. Consequently, this gives us the opportunity to compare contrastive loss with CPL more accurately and to use ${train}_{cpl}$ dataset.

The equation of the adapted contrastive loss with the notation identical to Equation \eqref{eq:2}:
\begin{equation}
    \begin{aligned}
    L_{\text{contrast}} =\ &\frac{1}{m}\sum_{i=1}^{m}(\left[\mathrm{dist}(o_{k_i},s_{i}) \right]_{+} \cdot l_i + \\ &\left[ \zeta - \mathrm{dist}(o_{k_i},s_{i}) \right]_{+} \cdot (1 - l_i))
    \end{aligned}
\end{equation}
Total loss is calculated similarly to Equation \eqref{eq:1}:
\begin{equation}
    L = L_{\text{contrast}} \cdot \lambda + L_{\text{CE}}
\end{equation}
\subsubsection{Dataset and implementation}
For training and evaluation, the same datasets ${train}_{cpl}$ and ${test}_{cpl}$ are used. The only difference is that unique class identifiers ${k}_{i}$ are not used. The implementation ${model}_{contrast}$ follows \autoref{sec:3.2} and differs only in the contrastive loss implementation inside the forward pass.
\subsubsection{Experimentation}
To evaluate adapted contrastive loss we conducted 42 experiments utilizing grid search for hyperparameters $\zeta \in [0.03, 0.18]$ with a step 0.03 and $\lambda \in [1.00, 1.30]$ with a step 0.05, following the same settings as in \autoref{sec:3.3}. The rationale for choosing the interval of $\lambda$ is similar to CPL, while the span for the margin $\zeta$ is based on typical margin values for contrastive loss and the fact that it should be positive.

\begin{figure*}[h]
    \centering
    \includegraphics[width=0.75\linewidth]{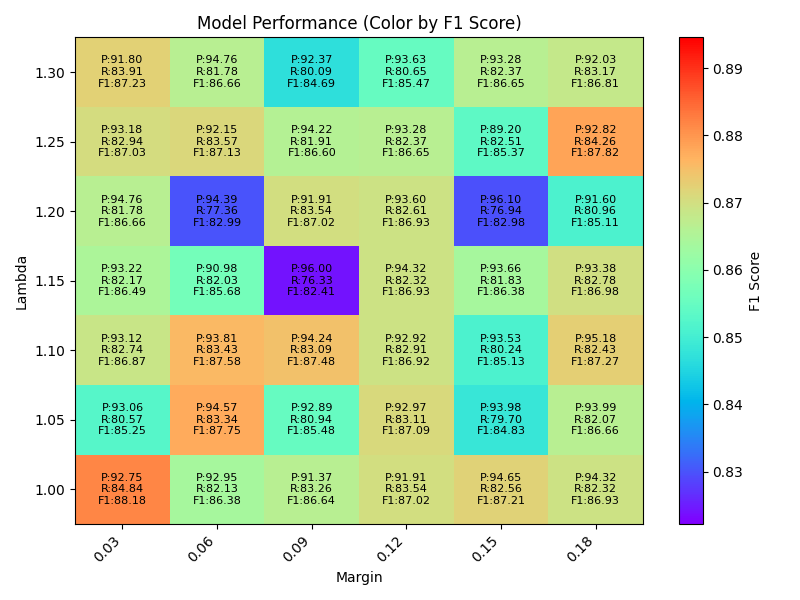}
    \caption{Performance matrix for the selected range of $\lambda$ and $\zeta$ of the $model_{contrast}$ fine-tuned on the Java dataset. Contrastive loss demonstrates lower highest and median F1-scores than Cluster Purge Loss does.}
    \vspace{-25mm}
    \label{fig:matrix_contrastive}
\end{figure*}

The results for all combinations of $\lambda$ and $\zeta$ are demonstrated in \autoref{fig:matrix_contrastive}. The best result in terms of F1-score is (P: \textbf{92.75\%}, R: \textbf{84.84\%}, F1: \textbf{88.18\%}). The best outcome of ${model}_{contrast}$ shows \textbf{0.96 pp} increase in F1-score compared to ${model}_{base}$. However, in turn, ${model}_{CPL}$ outperforms ${model}_{contrast}$ with a difference of \textbf{1.28 pp}. The median F1-score in the explored hyperparameter range is 86.65\% which is 0.87 pp less than 87.52\% shown by CPL(p < 0.0001).

Thus, we can conclude that although adding contrastive loss as a second term in joint formulation with cross-entropy loss improved the results compared to the baseline, Cluster Purge Loss achieves greater gains in all metrics.

\section {Generalizability study}
\label{sec:generalization_appendix}
The main dataset used in this study was the Java mutant dataset which was derived by \citet{tian_large_2024} from MutantBench\citep{noauthor_mutantbench:_2021}. It was chosen for the following reasons:
\begin{enumerate}
    \item Using this dataset allows for direct comparison with the results obtained for UniXCoder by Tian et al.
    \item MutantBench is the largest mutant dataset which is aggregated from previous public ones to date.
    \item Tian dataset was preprocessed to include method-level mutants and be suitable to be used as input for UniXCoder.
\end{enumerate}
The domain of equivalent mutant detection suffers from a scarcity of high-quality datasets. To evaluate how CPL generalizes to other datasets, especially consisting of mutants written in other languages, the only option is to use the second dataset from MutantBench compiled of mutants in C language. However, it is significantly smaller than the Java dataset and requires additional preprocessing.

\subsection{C dataset preprocessing}
MutantBench consists of program-level mutants - each mutant of an original program is another program modified with mutation operators. However, all mutation operators used in MutantBench are applied at the method level. For each program-level mutant we identified modified methods and isolated them, subsequently joining with their original counterparts, thereby constructing a method-level pairs of mutants, each of which was assigned a unique identifier. After converting the dataset to a structure consistent with that of the base Java dataset, we obtained 1099 mutants, of which 918 were equivalent and 181 non-equivalent. Mutants were split 50/50 into ${train^C}_{base}$ and ${test^C}_{base}$ maintaining the same ratio between equivalents and non-equivalents. Finally, to obtain  ${train^C}_{cpl}$ and ${test^C}_{cpl}$, preprocessing steps described in \autoref{sec:3.1} are applied. It is worth noting that in this dataset, equivalent mutants make up 83\%, while in the Java mutant dataset their share was only 15\%.

\subsection{Experiments}
\subsubsection{Cluster Purge Loss and the base model}
To evaluate ${model}_{cpl}$ on ${train^C}_{base}$, we fine-tuned it and conducted 56 experiments on exploring the same hyperparameter space provided in \autoref{sec:3.3}. The best result in terms of F1-score is (P: 97.30\%, R: 95.52\%, F1: 96.38\%) at $\lambda$ = 1.3 and $\zeta$ = -0.01. Taking into account that the metrics of ${model}_{base}$ fine-tuned on this dataset is (P: 94.15\%, R: 95.37\%, F1: 94.74\%), the absolute gain is (P: \textbf{3.15 pp}, R: \textbf{0.15 pp}, F1: \textbf{1.64 pp}).

\subsubsection{Contrastive Loss}
To assess ${model}_{contrast}$ 42 experiments were carried out in the similar fashion and with the same range of hyperparameters as described in \autoref{sec:experiments_appendix}. The best result of the contrastive setup in terms of F1-score is (P: \textbf{96.88\%}, R: \textbf{93.75\%}, F1: \textbf{95.22\%}) at $\lambda$ = 1.05 and $\zeta$ = 0.09. Thus, ${model}_{contrast}$ demonstrates \textbf{0.48 pp} increase in comparison with ${model}_{base}$, but ${model}_{cpl}$ still outperforms it with a difference of \textbf{1.16 pp}.

\begin{table}[h]
    \centering
    \resizebox{0.9\columnwidth}{!}{
      \begin{tabular}{|lccc|}
         \hline
         \textbf{Technique} & \textbf{Precision} & \textbf{Recall} & \textbf{F1-score} \\
         \hline\hline
         ${model}_{CPL}$ & \textbf{97.30\%} & \textbf{95.52\%} & \textbf{96.38\%} \\
         ${model}_{contrast}$ & 96.88\%, & 93.75\% & 95.22\% \\
         ${model}_{base}$ & 94.15\% & 95.37\% & 94.74\%  \\
         \hline
      \end{tabular}
    }
    \caption{Performance of different models on  the C mutant dataset}
    \label{tab:comparison}
    \vspace{-5mm}
\end{table}

\subsection{Conclusions}
After conducting experiments using the C mutant dataset, we observed that the models exhibited the same ranking in terms of precision, recall and F1-score as was obtained as a result of experiments on the main Java mutants dataset: Contrastive loss yields higher results than the baseline model, but our proposed Cluster Purge Loss model, in turn, outperforms the Contrastive loss model in all metrics.
Thus, our findings regarding CPL performance carry over to a distinct dataset of C‑based mutants, which features a substantially different proportion of equivalents. We consider this a strong indicator of the generalizability of our proposed approach.

\section{Scientific artifacts usage}
Pre-trained UniXCoder model, ${train}_{base}$, ${test}_{base}$, ${model}_{base}$ are obtained from \url{https://github.com/tianzhaotju/EMD} where the replication package for \citet{tian_large_2024} was released. It was sanctioned for replication, future research, and practical use, which we consider our usage to fall under.
\section{Computational budget}
Each of the 98 reported experiments aimed at hyperparameter search for CPL and contrastive loss on the Java mutant dataset required approximately 2.25 GPU-hours on a single NVIDIA RTX 4060, yielding a computational cost of around 214 hours. The same set of 98 experiments conducted on the C mutant dataset added another 137 hours at 1.4 hours per experiment. The total computational cost thus equals 351 hours.

\vfill\break
\onecolumn
\section{Embeddings visualization}
\begin{figure}[h]
    \centering
    \includegraphics[width=0.9\linewidth]{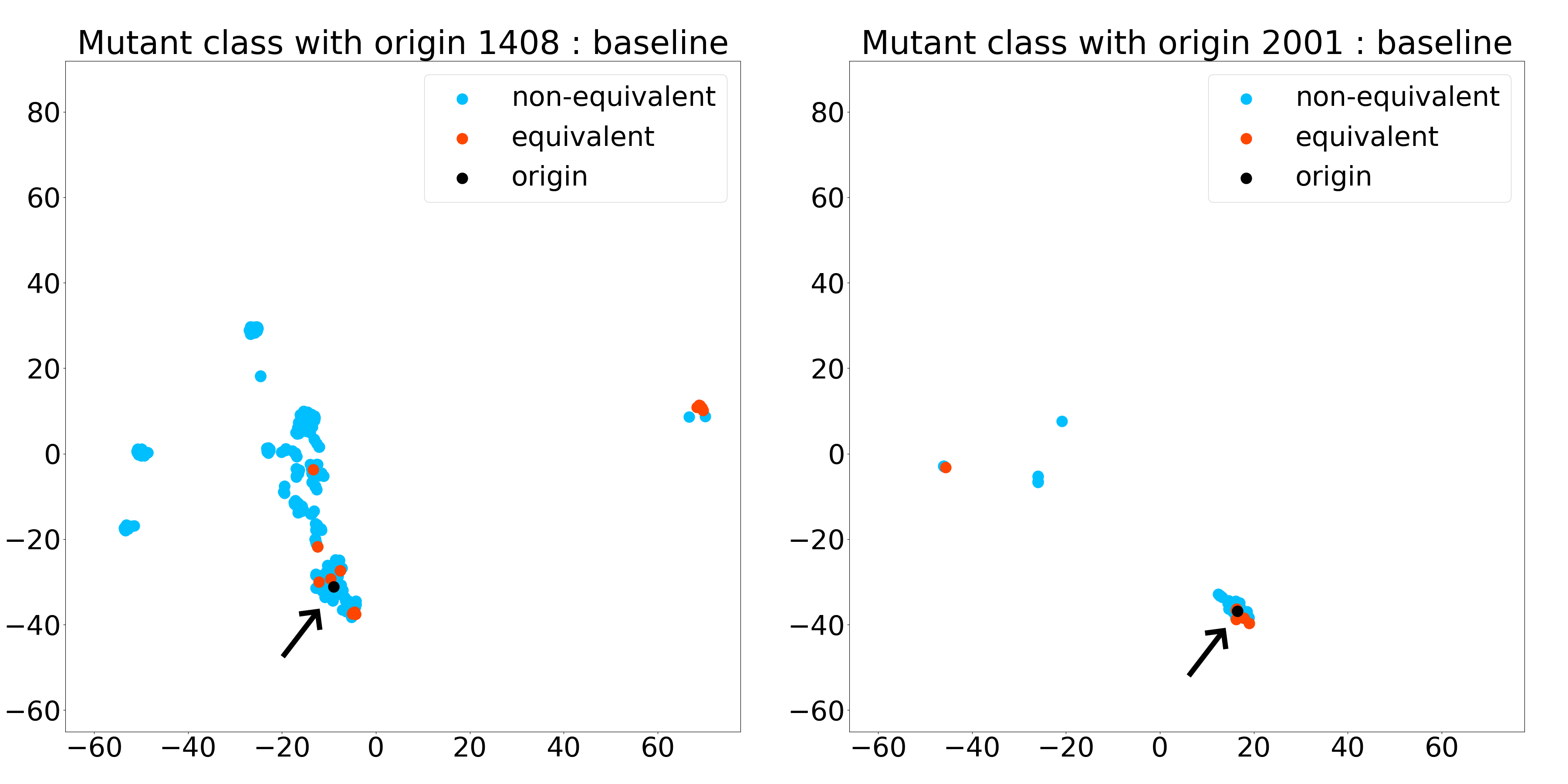}
    \caption{Baseline embeddings for classes with origins 1408 and 2001}
    \label{fig:baseline_embeddings}
\end{figure}
\begin{figure}[h]
    \centering
    \includegraphics[width=0.9\linewidth]{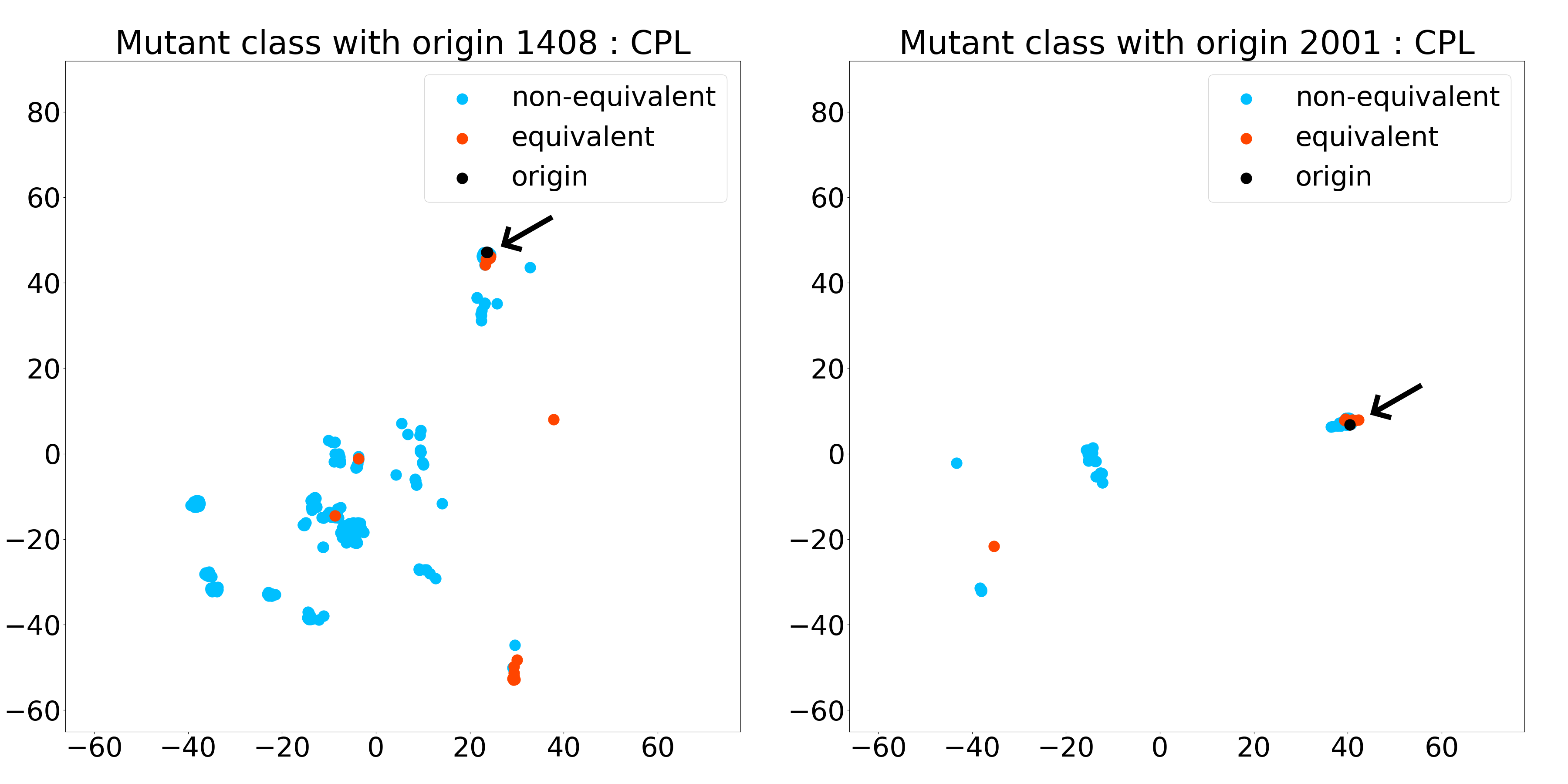}
    \caption{Embeddings after Cross-Entropy and Cluster Purge Loss for classes with origins 1408 and 2001}
    \vspace{-5mm}
    \label{fig:embeddings_cpl}
\end{figure}
\vfill\break

\section{Mutation testing}
\label{sec:mutesting_appendix}
\begin{figure*}[h]
    \centering
    \includegraphics[width=0.9\linewidth]{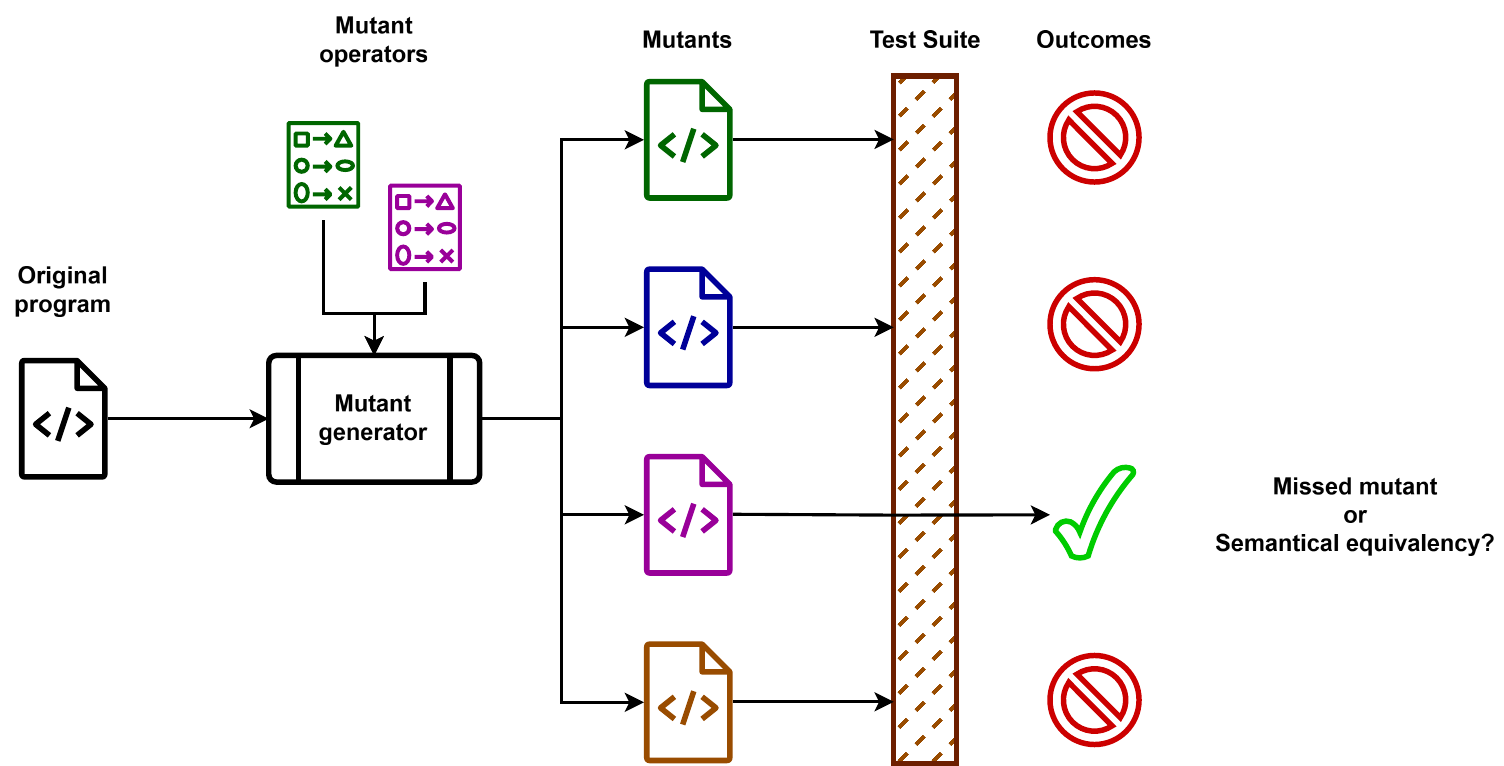}
    \caption{The process of mutation testing of the test suite for the original program. Equivalent mutants don't let make conclusions about the true mutation score.}
    \label{fig:mutesting}
\end{figure*}

\section{Mutants examples}
\label{sec:examples_appendix}
\begin{table*}[ht]
\centering
\begin{tabular}{|p{0.3\linewidth}|p{0.3\linewidth}|p{0.3\linewidth}|}
\hline
\textbf{Origin} & \textbf{Non-equivalent Mutant} & \textbf{Equivalent Mutant} \\
\hline
\begin{lstlisting}[language=Java, ]
int binSearch(int arr[], int x) {
int l = 0;
int h = arr.length - 1;
while (l <= h) {
    int mid = l + (h - l) / 2;
    if (arr[mid] == x)
        return mid;
    if (arr[mid] < x)
        l = mid + 1;
    else
        h = mid - 1;
}
return -1;
}
\end{lstlisting}
&
\begin{lstlisting}[language=Java, numbers=none]
int binSearch(int arr[], int x) {
int l = 0;
int h = arr.length - 1;
while (l <= h) {
    int mid = l + (h - l) / 2;
    if (arr[mid++] == x)
        return mid;
    if (arr[mid] < x)
        l = mid + 1;
    else
        h = mid - 1;
}
return -1;
}
\end{lstlisting}
&
\begin{lstlisting}[language=Java, numbers=none]
int binSearch(int arr[], int x) {
int l = 0;
int h = arr.length - 1;
while (l <= h) {
    int mid = l + (h - l) / 2;
    if (arr[mid] == x)
        return mid++;
    if (arr[mid] < x)
        l = mid + 1;
    else
        h = mid - 1;
}
return -1;
}
\end{lstlisting}
\\
\hline
\end{tabular}
\caption{Examples of code mutants. The first column shows the origin method intended to perform binary search on array \textit{arr} to find \textit{x}. The second column is a non-equivalent mutant created by applying Unary Operator Insertion (UIO) to the line 6. Post-increment affects the return statement inside if clause resulting in returning the wrong value.The third column shows an equivalent mutant produced by applying UIO to the line 7. In this case post-increment doesn't influence behaviour as the method execution ends.}
\label{tab:mutants}
\end{table*}
\break
\section{Model performance matrix}
\label{sec:matrix_appendix}
\begin{figure*}[h]
    \centering
    \includegraphics[width=0.75\linewidth]{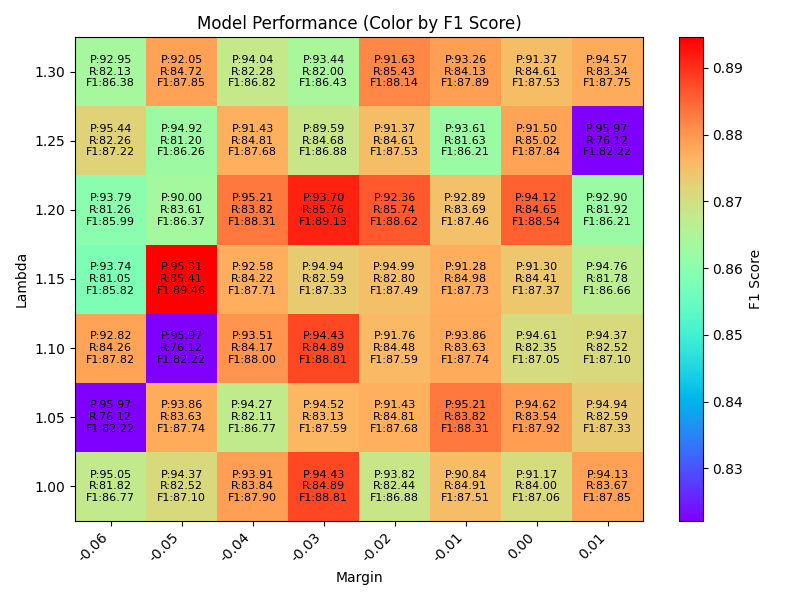}
    \caption{Model performance matrix for different values of $\lambda$ and $\zeta$ presenting Precision(P), Recall(R), F1-score(F1) of the ${model}_{CPL}$ fine-tuned on the Java dataset. Also, it can be observed that the matrix of metrics is heterogeneous, that can be attributed to the non-linear nature of interaction between $\lambda$ and $\zeta$.}
    \label{fig:matrix}
\end{figure*}




\end{document}